\documentclass[floatsintext]{apa7}

\usepackage{lipsum}

\usepackage[american]{babel}

\usepackage{csquotes}
\usepackage[style=apa,sortcites=true,sorting=nyt,backend=biber]{biblatex}
\DeclareLanguageMapping{american}{american-apa}
\title{Proverbs As A Window On Collective Prescriptions}

\shorttitle{Proverbs}

\author{Tomás Lejarraga$^{1,2}$, Dirk U. Wulff$^{3,2}$, and Ralph Hertwig$^{2}$}
\affiliation{$^1$Universitat de les Illes Balears\\$^2$Max Planck Institute for Human Development\\$^3$Vienna University of Economics and Business}

\leftheader{Lejarraga, Wulff, Hertwig}

\abstract{Proverbs, as repositories of traditional wisdom, have informed human reasoning for millennia. Yet the collective prescriptions they encode have remained largely unquantified. We introduce a computational framework to analyze nearly 50,000 English-language proverbs. Using large language models, validated against human judgments, we map the topical landscape of proverbs, assess their prescriptive power, and quantify their recommendations on key dimensions of human judgment and behavior. We find that most proverbs are prescriptive, offering guidance in domains under individual control, such as preparation and prudent action, whereas proverbs about uncontrollable circumstances, such as fate and mortality, tend to describe rather than prescribe. When prescriptive proverbs take a consistent stance, they favor caution over risk, long-term over short-term outcomes, honesty over deception, and kindness over malice, and show a moderate tendency toward suspicion over trust. On dimensions involving genuine trade-offs, especially autonomy versus compliance, proverbs provide contradicting advice rather than a single recommendation. This analysis provides a quantitative map of the collective prescriptions in a culture's proverbial wisdom, revealing a core of prudent, honest, and forward-looking guidance alongside varied advice where circumstances dictate the best course of action.}


\authornote{

\addORCIDlink{Tomás Lejarraga}{0000-0001-6135-2243}
\addORCIDlink{Dirk U. Wulff}{0000-0002-4008-8022}
\addORCIDlink{Ralph Hertwig}{0000-0002-9908-9556}

Correspondence concerning this article should be addressed to Tomás Lejarraga, Universitat de les Illes Balears, Ctra. de Valldemossa km 7.5, Palma, Spain. E-mail: tomas.lejarraga@uib.eu}

\begin{document}

\maketitle

\section{Introduction}

“The wisdom of proverbs has guided people in their social interactions for thousands of years throughout the world” \parencite[p. xi]{mieder2004proverbs}. The time-honored lineage, however, does not mean that they are on their way to extinction in a modern and complex world. “Another day, another dollar” or “Garbage in, garbage out” from the world of business and computers are of more recent vintage. According to Mieder \citeyear{mieder2004proverbs}, proverbs are “indeed alive and well, and as sapient nuggets, they continue to play a significant role in the modern age” \parencite[p. xi]{mieder2004proverbs}. Proverbs have indeed been extensively investigated from different disciplinary angles such as semiotics, linguistics, cognitive science, literature, media studies and their importance has even given rise to a field of proverb studies, called \textit{paremiology} \parencite[see][]{hrisztova2015introduction}. It is devoted to collecting and classifying proverbs, and examine the form, structure, style, content, function, meaning, and value of proverbs. 

Proverbs' "economised form helps the speaker to remember and recognise them easily. ... [they] allow their users to comment on, standardise and evaluate new situations with the help of known social clichés" \parencite[p. 182]{lewandowska20147}. Proverbs serve myriad functions, including recommending an action, reconstructing experiences, explaining and reasoning about the past, offering instructions and warnings for the future, as well as encouraging and amusing others \parencite{lauhakangas2007use}. As diverse as their functions are the subjects in which proverbs are invoked for guidance and sense-making, including practical knowledge of nature, faith and existence, sense of proportion, morality, social interaction, communication, agreements and norms, sense of duty, coping and learning, and sense of time (Norrik, 2014). Employed and likely created by individual speakers for communicative purposes \parencite{mieder2014origin}, their use and success require the acceptance by the people at large \parencite{firth1926proverbs}. Therefore, they also embody judgments and expressions of traditional and contemporary views and beliefs representative of a \textit{conscience linguistique collective} (‘collective consciousness’) \parencite[p. 100]{anscombre1994proverbes}. By extension, proverbs provide a window on collective prescriptions.

\begin{figure*}[htbp!]
\centering
\includegraphics[width=.95\textwidth]{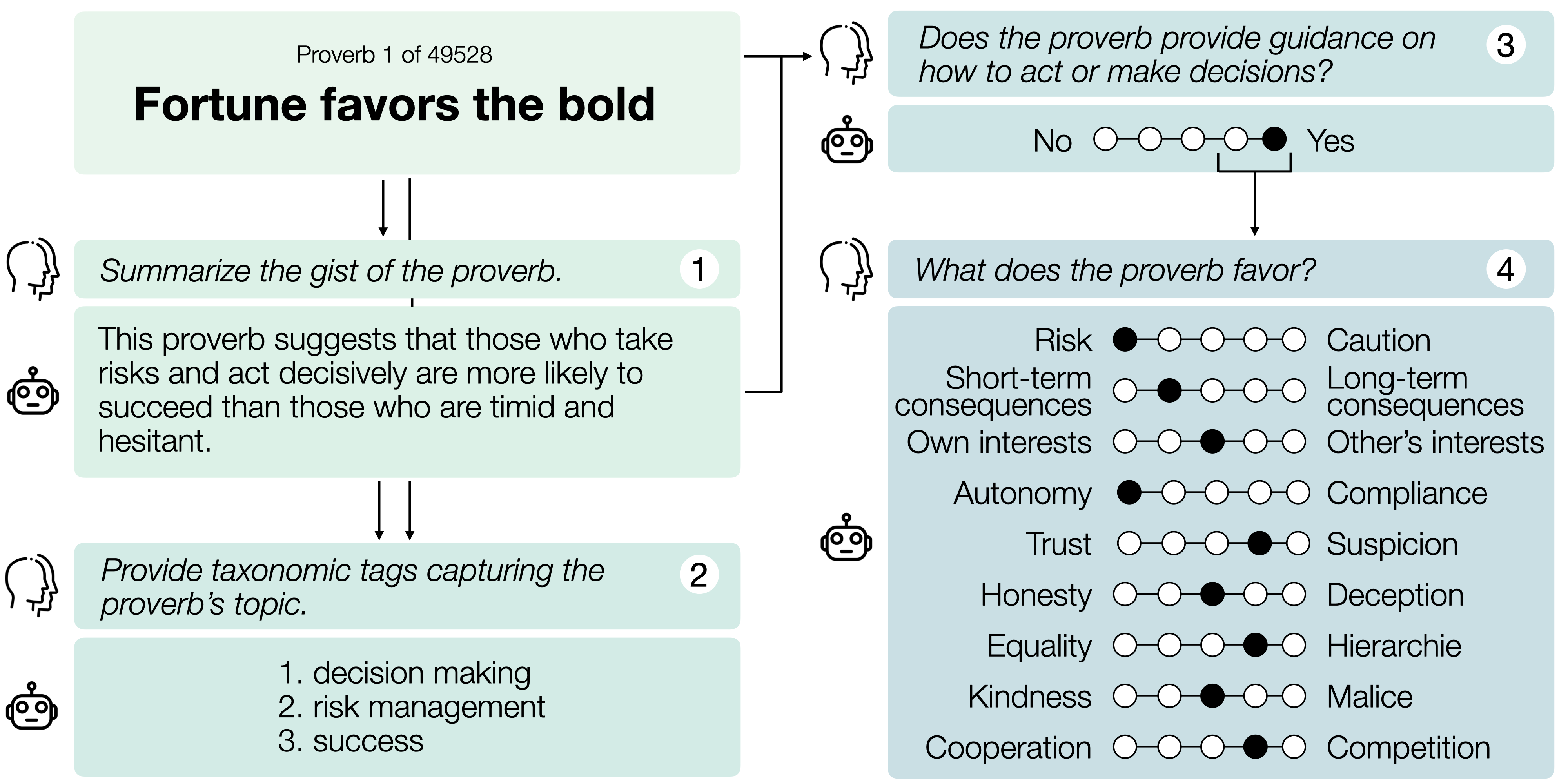}
\caption{Procedure to retrieve proverb labels, prescriptive capacity, and rating of preferences and values.}
\label{fig:approach}
\end{figure*}

It is this potential of proverbs that we study here. To open a window onto the collective consciousness expressed through proverbs, we introduce an analytical-computational framework for characterizing and quantifying their content (see Figure\ref{fig:approach}). We address three research questions. First, using a sample of 49,528 English-language proverbs and large language models, we classify the topics addressed by each proverb based on both its wording and a description of its gist. This analysis identifies the situations represented in proverbs and their relative prevalence. Second, we examine the extent to which proverbs \textit{prescribe} rather than describe how individuals should act or decide. This distinction parallels prominent accounts of social norms that differentiate social regularities (what people do'') from normative guidance (what people ought to do'') \parencite{cialdini1990focus,bicchieri2005grammar}. Third, we analyze the direction of these prescriptions across nine fundamental dimensions of human judgment and behavior. Three are central to economic research because they "affect such a broad range of economic decisions .... and [afford] the opportunity to better explain economic behavior" \parencite[p. 1935]{falk2023preference}: risk preference (i.e., recommendation of taking risks or exercising caution), intertemporal preference (i.e., recommendation of actions favoring short-term or long-term consequences) and social preference (i.e., recommendation of own-interest or other-regarding behaviors). The remaining dimensions are compliance vs. autonomy \parencite[e.g.,][]{taylor2010autonomy, gruen1976autonomy}, cooperation vs. competition \parencite[e.g.,][]{gachter2010culture}, equality vs. hierarchy \parencite[e.g.,][]{hofstede1984culture,rokeach1967rokeach}, honesty vs. deception \parencite[e.g.,][]{levine2014truth}, kindness vs. malice \parencite[e.g.,][]{baron2011zero}, and trust vs. suspicion \parencite[e.g.,][]{inglehart2006mapping}. For each dimension, we assess which pole is favored by prescriptive proverbs.

Of the seven collection of proverbs used here, five focus on American or British proverbs, whereas the other two are multicultural or global in scope. We analyze the collections in aggregate and also report results separately for each collection, allowing us to assess the extent to which the wisdom expressed in proverbs is shared across cultural traditions rather than specific to English-speaking contexts, even though this is not our focus here.

\section{Results}

\subsection{Using LLMs to evaluate the meaning of proverbs}

To map the collective consciousness reflected in proverbs, we first charted the semantic landscape of our corpus using LLMs. Where previous studies were limited to the manual examination of small samples, our computational approach permits a systematic analysis of a large corpus of proverbial record.

\begin{figure*}[htbp!]
\centering
\includegraphics[width=.95\textwidth]{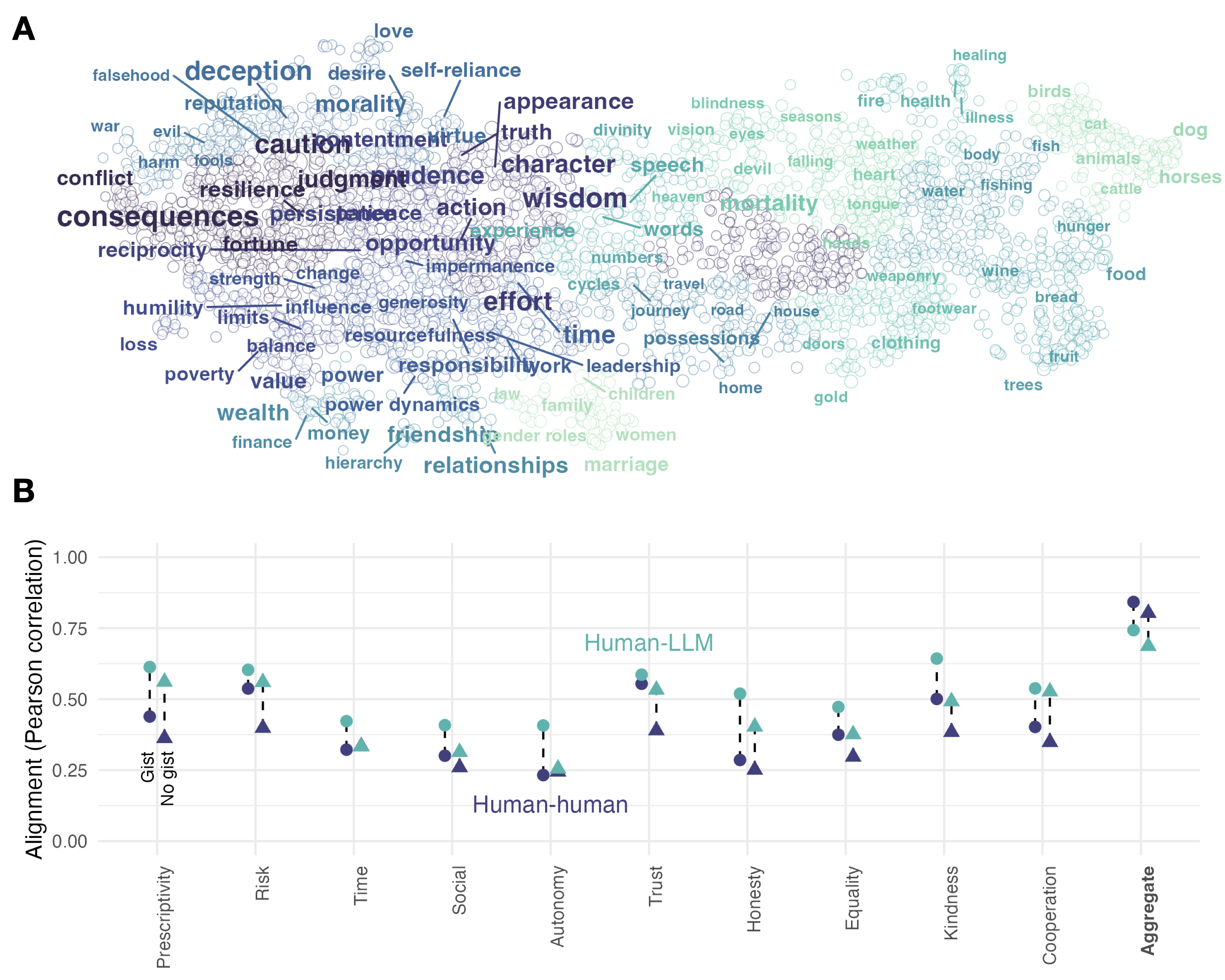}
\caption{Topical landscape of proverbs and validation of model ratings. (A) Two-dimensional projection of the semantic space of proverb topics. Each circle represents a topic tag, and topics appearing closer together are semantically similar. Labels mark the most frequent topics, with font size proportional to topic frequency and colors indicating clusters of similar topics. (B) Alignment of proverb ratings, measured as Pearson correlations, for each rated dimension and for the aggregate analysis across the nine dimensions of judgment and behavior. Green points show the alignment between the average human rating and the model's ratings, and purple points show the alignment among human raters. Circles and triangles show ratings obtained with and without the model-generated gist explanation, respectively.}
\label{fig:map}
\end{figure*}

Figure \ref{fig:map} shows the topical landscape of proverbs, visualized as a semantic map in which proximity indicates thematic similarity. The most frequent topic is, fittingly, wisdom itself (n = 1,465), suggesting that proverbs often reflect on the very knowledge they are meant to transmit. Several of the most frequent topics concern foresight and careful action, including consequences (n = 1,377), caution (n = 1,354), and prudence (n = 1,127), which foreshadows the prescriptions we identify in the following. Proverbs also frequently address deception (n = 1,167), the social world, as in relationships and friendship, and even existential themes such as mortality. This topical profile shows that proverbs focus on the perennial challenges of judgment and social life.

To validate our computational approach, we compared the model's ratings of proverbs against the judgments of human raters. At the level of individual proverbs, the model's ratings agreed more strongly with the average human judgment than human raters agreed with each other, and this held for every dimension we assessed. Agreement was higher when raters were shown the model-generated explanation of a proverb's gist, but the pattern of results was unchanged without it. When ratings were aggregated across proverbs to characterize the nine key dimensions, the correspondence between model and human judgments was high (r = .74). These results indicate that the model captures the human understanding of proverbs, supporting its use for analyzing proverb content at a scale unattainable through manual coding.

\subsection{Scoring the prescriptivity of proverbs}

\begin{figure*}[htbp!]
\centering
\includegraphics[width=.95\textwidth]{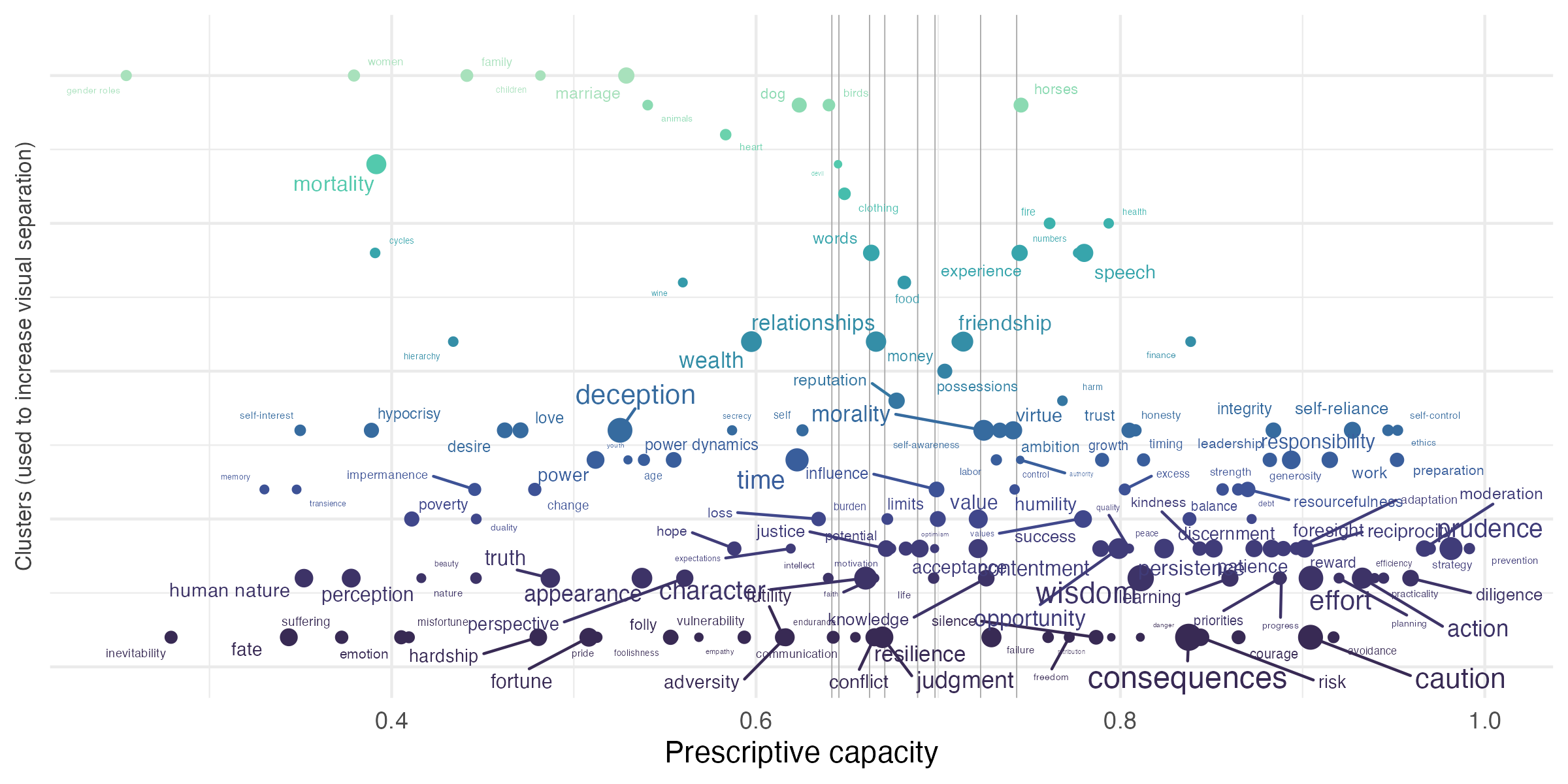}
\caption{Average prescriptive capacity of proverbs by topic. Higher values indicate that proverbs within a topic provide more guidance on how to act or decide. Point size is proportional to the number of proverbs in a topic, and colors correspond to the topic clusters shown in Figure \ref{fig:map}. The vertical axis separates clusters for readability and carries no meaning. Grey vertical lines show averages by collections.}
\label{fig:prescriptivity}
\end{figure*}

Proverbs can guide behavior, as in ``Measure twice, cut once;'' but they can also simply describe the world, as in ``Every rose has its thorn.'' To quantify these two functions, we prompted an LLM to rate the prescriptive capacity of each proverb, that is, the extent to which it provides guidance on how to act or decide. Figure \ref{fig:prescriptivity} shows the average prescriptive capacity of proverbs for each topic.

We found that most proverbs prescribe. The LLM rated 67.1\% of proverbs as prescriptive, and human raters classified an even larger share as prescriptive (73.8\%). Prescription rather than description is thus the rule in the proverbial record.

Prescriptive capacity nevertheless varied systematically across topics. Proverbs were most prescriptive in domains of deliberate action, including preparation, strategy, prudence, prevention, caution, and moderation. These are domains in which behavior is under the individual's control and clear guidance is valuable. In contrast, prescriptive capacity was lowest for topics concerning circumstances that lie largely beyond individual control, such as fate, inevitability, mortality, human nature, and suffering. Proverbs in these domains tend to describe and make sense of the human condition rather than direct behavior. Analyses by collection, shown by the grey vertical lines, revealed no substantial differences in prescriptivity across the seven sources. This pattern suggests a division of labor in proverbial wisdom. Where action is possible, proverbs tend to take on a more prescriptive role, as in ``Make hay while the sun shines''; where it is not, they tend to console and explain, as in ``What will be, will be''. Even largely descriptive proverbs, however, may carry implicit guidance, because statements about what is often convey implications about what one ought to do \parencite{bicchieri2005grammar}.

\subsection{Evaluating the polarity of proverb prescriptions}

A fascinating characteristic of proverbs is their ability to advocate opposing behaviors, a quality that has long intrigued scholars \parencite{breton1631crossing}. In the domain of risk taking, for instance, we find both ``Fortune favors the bold'' and ``Look before you leap.'' Similarly, ``Many hands make light work'' promotes joint effort, whereas ``Too many cooks spoil the broth'' warns against it. Despite the prominence of such contradictions, it remains unclear whether they are a general feature of proverbial wisdom or are concentrated on particular dimensions of judgment and behavior. Some dimensions may give rise to competing forms of advice, whereas others may be associated with a clear consensus.

To address this question, we characterized the recommendations of prescriptive proverbs, that is, proverbs rated as providing guidance (``yes'' or ``strong yes''), on each dimension using two measures, computed separately for LLM and human ratings. Polarity captures the balance of the recommendations, ranging from 0, when all recommendations favor the same pole, to 1, when recommendations are evenly split between the poles. Extremity captures how often proverbs take a stance on a dimension, measured as the proportion of ratings outside the midpoint category. 

\begin{figure*}[htbp!]
\centering
\includegraphics[width=.95\textwidth]{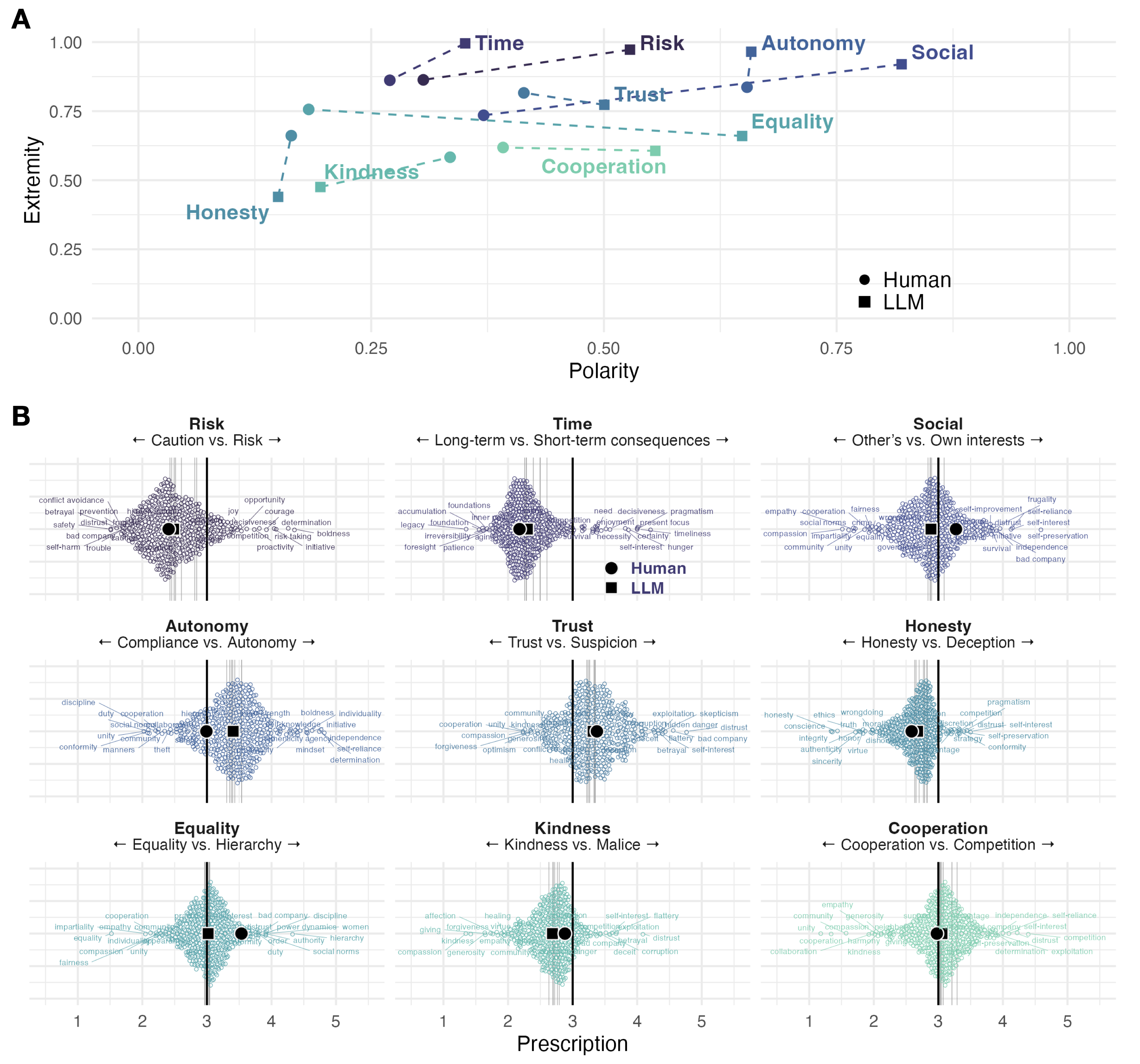}
\caption{Recommendations of prescriptive proverbs (rated ``yes'' or ``strong yes'' on prescriptivity) on the nine dimensions of human judgment and behavior. (A) Extremity and polarity of proverb recommendations, based on human ratings (circles) and LLM ratings (squares). Extremity is the proportion of prescriptive proverbs rated outside the midpoint category. Polarity captures the balance of recommendations, ranging from 0, when all recommendations favor the same pole, to 1, when recommendations are evenly split between the poles. Dashed lines connect the two ratings of the same dimension. (B) Average recommendations by dimension. Each panel shows one dimension, with arrows indicating the poles. Small points show the average rating of proverbs within a topic, and labels mark the topics leaning most strongly toward either pole. Large points show the overall average of proverb ratings by human raters (circles) and by the LLM (squares). The vertical line marks indifference between the two poles; averages to its left indicate a leaning toward the left pole and averages to its right a leaning toward the right pole. Grey vertical lines show averages by collections.}
\label{fig:polarity_preferences}
\end{figure*}

Figure \ref{fig:polarity_preferences}A shows that contradiction is a selective rather than a general feature of proverbial wisdom. LLM and human ratings converge in identifying autonomy versus compliance as a highly contested dimension, with recommendations split markedly between the poles (polarity of about .65 for both). For equality and social interests, in contrast, the LLM perceived substantially more contradiction than human raters, who saw proverbs as speaking with a more unified voice. In contrast, both LLM and human raters consistently viewed the proverbs as prescribing kindness over malice and honesty over deception, with little contradictory proverbs. We return to the direction of these recommendations in the next section.

In some dimensions, proverbs are more extreme than in other dimensions. According to both LLM and human ratings, proverbs are more extreme for the classic trade-offs of decision making, namely time, risk, autonomy, and own versus others' interests, suggesting that a stance on these dimensions is nearly inseparable from prescriptive advice. In contrast, proverbs about honesty and kindness tend to be less extreme. Moreover, polarity was strikingly consistent across collections, as indicated by the vertical grey lines, suggesting that the prescriptions expressed in proverbs share common features across cultural traditions.

Overall, this pattern suggests that proverbial wisdom speaks with one voice on moral virtues, such as honesty and kindness, but preserves competing counsel where interests and situations genuinely conflict, most clearly on the tension between autonomy and compliance. Rather than a set of dogmatic rules, proverbs seem to be a repertoire of context-dependent wisdom, offering firm norms where consensus exists and alternative strategies where it does not.

\subsection{Charting collective prescriptions}

Having established that most proverbs prescribe, we can now ask what they prescribe. Figure \ref{fig:polarity_preferences}B maps the average recommendation on each of the nine dimensions of judgment and behavior among prescriptive proverbs, as rated by the LLM and by human judges. We observe broadly similar patterns across the seven proverb collections (as indicated by the vertical gray lines), indicating that the aggregate results are not attributable to a single collection.

Proverbial wisdom takes a clear side on four dimensions, and the model and human raters agree on all of them. Proverbs endorse caution over risk taking, as in ``Look before you leap,'' long-term over short-term consequences, as in ``Don't put all your eggs in one basket,'' honesty over deception, as in ``Honesty is the best policy,'' and kindness over malice, as in ``Good to forgive, best to forget.'' The endorsement of long-term thinking showed the clearest and most consistent advice. These prescriptions closely track empirical findings on human preferences, including widespread risk aversion \parencite{bernoulli1738specimen,falk2023preference}, the economic benefits associated with long-term orientation \parencite{falk2023preference,becker1997accounting,lawrance1991poverty}, and the observation that people are often truthful and generous even at a cost to themselves \parencite{gneezy2005deception,abeler2019preferences,fehr2003nature}. Proverbs may help sustain these tendencies by transmitting the corresponding norms across generations and cultural contexts.

A clear and surprising pattern also emerged for trust. Proverbs lean toward suspicion rather than trust, as in ``Trust, but verify.'' This tendency was evident across collections, suggesting that proverbial traditions from different cultural sources share a concern with the risks of misplaced trust. This may reflect a basic asymmetry: the benefits of trust are often small and gradual, whereas its failures can be immediate and costly. Trust enables cooperation and exchange, but it also leaves people vulnerable to exploitation \parencite{berg1995trust,knack1997does,putnam1993making}. Proverbial wisdom appears to place greater weight on avoiding exploitation.\footnote{This pattern was not observed in Figure 4A, where the polarity measure reflects only the direction of the prescription and does not incorporate its strength.}

The remaining dimensions show little overall consensus for one side or the other. Recommendations concerning cooperation versus competition, own versus others' interests, autonomy versus compliance, and equality versus hierarchy average close to indifference. This does not reflect an absence of advice. On the contrary, these are the dimensions that generated highly contradictory prescriptions. Their neutral averages arise because opposing messages coexist within the same dimension and largely offset one another. This coexistence of opposing prescriptions was observed across collections, indicating that it is not merely an artifact of combining culturally distinct sources. Rather, proverbial traditions appear to preserve competing counsel where the best course of action depends on circumstances.

These contested dimensions are also where model and human ratings diverge most. Human raters perceived proverbs as leaning toward hierarchy and toward own-interest, whereas the model rated them as neutral or mildly favoring others' interests, and the model perceived a leaning toward autonomy that human raters did not perceive. Given the strong overall agreement between model and human ratings, these divergences are the exception rather than the rule, but they suggest caution in interpreting the precise position of proverbs on the contested dimensions.

Taken together, the proverbs reveal a coherent set of priorities that is broadly consistent across collections. They favor caution over risk, patience over immediacy, honesty over deception, and kindness over malice. At the same time, they are skeptical of misplaced trust. On questions of social organization, however, they are less decisive. Cooperation and competition, autonomy and compliance, equality and hierarchy all attract competing advice. 

\section{Discussion}

This work makes a significant contribution by moving the study of proverbs from qualitative interpretation to quantitative cultural analytics. The use of LLMs allows us to analyze cultural products at an unprecedented scale \parencite{hussain2024tutorial, binz2025should}, revealing systematic patterns in the repository of proverbial wisdom. Our principal insight is that this wisdom is not random. Where proverbs take a consistent side, they advise caution, long-term thinking, honesty, and kindness, tempered by some skepticism of misplaced trust. This aggregate advice paints a portrait of a cultural toolkit designed to foster prudent behavior and reliable social relations. The recurrence of these patterns across collections is suggestive of cross-cultural consistency, although the predominantly American and British source base warrants caution. As cultural products, proverbs store and transmit this wisdom, reflecting the values that have emerged as adaptive over generations \parencite{weber1998cross}.

Equally revealing is where proverbial wisdom does not speak with one voice. On dimensions involving genuine and difficult trade-offs, most clearly the tension between autonomy and compliance, proverbs offer competing advice, and their recommendations balance each other out in the aggregate. Rather than a set of dogmatic rules, the proverbial record thus resembles a repertoire of context-dependent guides, offering firm norms where consensus exists and alternative strategies where the best course of action depends on circumstances. This may explain why the contradictions among proverbs, long noted by scholars \parencite{yankah1984proverbs}, coexist with their authority as vehicles of advice.

This study provides the groundwork for future research. An obvious extension is cross-cultural comparison. Such work could test whether documented differences in risk attitudes and social preferences across countries \parencite{falk2023preference} are mirrored in the prescriptions encoded in proverbial traditions, a hypothesis previously explored only with small-scale studies \parencite{weber1998cross}. Another testable hypothesis emerged in our analysis. It appears as if that proverbs from different cultural sources offer clear, consistent and culture-general guidance in domains where the costs and benefits are, ceteris paribus, relatively context-independent and established such as when they highlight the value of patience over immediacy (e.g., compound effects, information accumulation over time, cost of mistakes, better cooperation). In contrast, they preserve multiple, even opposite and, perhaps, culture-specific perspectives where the best course of action depends on the circumstances (e.g., autonomy vs. compliance). In studying these and other questions of culture-specificity or generality, the main challenge is attributing available and published proverbial corpora to cultures unambiguously--commonly corpora represent collections across cultures. 

Another future extension is to compare our topic taxonomy could with existing, manually crafted paremiological classifications to bridge computational and traditional humanities scholarship. The value of our framework lies in its capacity to scale up such analyses, transforming proverbs into a quantitative resource for tracking cultural values over time and across societies.

\subsection{Limitations}
Our analysis has limitations. Many proverbs are time-honored artifacts, reflecting the views of a past era. Proverbs like ``Don't put the cart before the horse'' highlight the historical context in which they emerged. This historical orientation is inherent to the study of proverbial tradition itself. For a saying to become established as a proverb, it must be retained across generations. As a result, proverb collections are particularly well suited to capturing culturally durable views, but are less likely to reflect ideas that have only recently emerged. Although new proverbs continually enter the lexicon to reflect modern life (``The internet never forgets''), our corpus is naturally biased toward views that have stood the test of time. Future work could incorporate sources of modern proverbs to capture the views of the current information age that may become proverbial in the future, such as ``Data is the new oil,'' which equates data with a critical economic resource. Together, these sources could help trace both the persistence of collective prescriptions and their evolution over time.

Another limitation concerns the type-token distinction. We analyzed each proverb as a unique type, irrespective of its prevalence in natural language (its token frequency). An alternative approach would be to analyze the frequency of each proverb's occurrence. However, because proverbs are more prevalent in spoken language and relatively rare in written corpora, such an analysis would require enormous volumes of text to yield reliable estimates.

A further limitation is that our corpus consists predominantly of English-language proverbs, although two of the seven sources comprise worldwide or multicultural collections. The main patterns were consistent across collections, suggesting some convergence across cultural traditions. Nevertheless, because collections do not map neatly onto cultures, research using non-English and culturally defined corpora is needed to establish cross-cultural generalizability. Let us finally highlight that there is still no consensus definition of what makes something a "proverb," as opposed to a proverbial phrase, idiom, maxim, slogan, or aphorism. Wolfgang Mieder—arguably the field's most prominent contemporary researcher—has spent decades refining definitions, yet the boundaries remain fuzzy. Is "garbage in, garbage out" a proverb? What about internet-era formulations like "pics or it didn't happen"? To the extent that proverbs and related constructs offer us a window onto collective wisdom, preferences, and cultural priors, we tend to prefer a wider rather than a narrower definition, and, therefore, we did not winnow down the corpora. 

\section{Methods}

\subsection{Identifying Proverbs}

We aggregated proverbs from seven published collections: \textit{A Dictionary of American Proverbs} \parencite{mieder1992dictionary}, \textit{The Dictionary of Modern Proverbs} \parencite{doyle2012dictionary}, \textit{Dictionary of Authentic American Proverbs} \parencite{mieder2021dictionary}, \textit{A Dictionary of Proverbs} \parencite{simpson2008dictionary}, \textit{Little Oxford Dictionary of Proverbs} \parencite{knowles2016little}, \textit{The Multicultural Dictionary of Proverbs} \parencite{cordry2015multicultural}, and \textit{The Routledge Book of World Proverbs} \parencite{stone2006routledge}. We extracted proverbs from the digitized collections using source-specific extraction rules, retaining main entries as well as variants listed in the collections. All entries were converted to lowercase, stripped of terminal punctuation and residual formatting, and deduplicated. Entries containing parenthetical variant insertions, which some collections use to indicate alternative wordings, were manually expanded into their full variants. This procedure yielded an initial list of 49,528 unique proverbs.

\textit{A Dictionary of American Proverbs} defines proverbs as concise statements of apparent truths in common circulation and includes traditional English-language and American-origin proverbs collected by the American Dialect Society between the 1940s and 1970s; it excludes proverbial expressions, comparisons, superstitions, wellerisms, and idioms \parencite{mieder1992dictionary}. \textit{The Dictionary of Modern Proverbs} focuses on proverbs originating in English since 1900 and, more broadly, on expressions current in modern usage. It defines ``true proverbs'' as complete or elliptical sentences that are formulaic but variable and express general observations or propositions, often figuratively; proverbial phrases, wellerisms, and sarcastic interrogatives are excluded \parencite{doyle2012dictionary}. \textit{The Dictionary of Authentic American Proverbs} provides no explicit definition but includes proverbs coined in North America between the seventeenth and twenty-first centuries \parencite{mieder2021dictionary}.

\textit{A Dictionary of Proverbs} defines a proverb as a traditional, concise saying that offers advice or presents a moral, and focuses on expressions used in Britain during the past two centuries, including some circulating throughout the English-speaking world \parencite{simpson2008dictionary}. The \textit{Little Oxford Dictionary of Proverbs} adopts a broader criterion, including proverbs and sayings from biblical, classical, global, and popular-cultural traditions, selected to represent diverse forms of actual usage, especially in advice and admonition \parencite{knowles2016little}.

Similarly, \textit{The Multicultural Dictionary of Proverbs} includes cross-cultural expressions of practical wisdom, as well as maxims understood as explicit or implicit rules of conduct \parencite{cordry2015multicultural}. Finally, \textit{The Routledge Book of World Proverbs} includes expressions that most folklorists and paremiologists would recognise as proverbs \parencite{stone2006routledge}.

Overall, these sources variously emphasize form, tradition, circulation, function, historical or geographical origin, and recognition by experts. We therefore adopt a compiler-based operational criterion: expressions are treated as proverbs when they are classified as such in at least one of these collections, subject to the additional selection criteria described in the main text.

\subsection{Obtaining Proverb Topics}

As depicted in Figure \ref{fig:approach}, we prompted Gemma-3-27B to generate up to three taxonomic tags for each proverb based on the proverb and its gist explanation. The system prompt instructed the model to act as ``a proverb scholar, skilled in assigning accurate and concise taxonomic tags to proverbs, considering both their general meaning and specific illustrative context.'' The user prompt presented the proverb and its gist and instructed the model to assign up to three tags of at most two words each, capturing the proverb's general meaning as well as, where applicable, its specific illustrative imagery, to briefly reason through its choices before finalizing them, and to return only the tags in a fixed format, which permitted automatic extraction. The complete prompt is available in the project repository. Extracted tags were converted to lowercase and cleaned of surrounding whitespace.

To consolidate variants of the same topic (e.g., singular and plural forms or close synonyms), we obtained a semantic vector representation (1,024 dimensions) for each unique tag using the Qwen3-Embedding-0.6B transformer \parencite{qwen3embedding}, embedding each tag with the instruction ``What is the meaning of this proverb category tag:''. We computed the angular similarity between all pairs of tags and grouped tags using complete-linkage hierarchical clustering, merging all pairs exceeding the top percentile of pairwise similarities. Each group of variants was represented by its most frequent tag, with a frequency equal to the summed frequencies of its members and a semantic vector equal to the average of its members' vectors.

To create the topic map (Figure \ref{fig:map}A), we projected the consolidated tag vectors into a two-dimensional space using PaCMAP \parencite{JMLR:v22:20-1061} with angular distance. To color the map, we grouped the tags into 20 clusters by applying complete-linkage hierarchical clustering to the two-dimensional coordinates. The map displays the most frequent tags overall and within each cluster.

\subsection{Rating Proverbs}

We used Gemma-3-27B \parencite[google/gemma-3-27b-it][]{team2025gemma} to rate each proverb on its prescriptive capacity and on the nine key dimensions of human judgment and behavior. For each rating, the model received the proverb together with its gist explanation. The system prompt instructed the model to act as ``an expert on proverbs and decision making who accurately assesses the meanings of proverbs.'' The user prompt then posed one of the ten assessment questions listed below, prefixed by ``Your task is to carry out the following assessment:'' and followed by the instruction ``To assess this, carefully examine both the proverb and its gist.'' The prompt further instructed the model to briefly reason through the assessment before finalizing it and to return its answer as one of five response options in a fixed format, which permitted automatic extraction of the ratings. The complete prompts are available in the project repository. The ten assessment questions and their response options (in parentheses) were:

``Does the proverb provide guidance on how to act or make decisions?'' (`strong no', `no', `indifferent', `yes', `strong yes')

``Does the proverb favor risk or caution?'' (`strongly favoring caution', `favoring caution', `indifferent', `favoring risk', `strongly favoring risk')

``Does the proverb favor short-term or long-term consequences?'' (`strongly favoring long-term consequences', `favoring long-term consequences', `indifferent', `favoring short-term consequences', `strongly favoring short-term consequences')

``Does the proverb favor your own or other people's interests?'' (`strongly favoring other people's interests', `favoring other people's interests', `indifferent', `favoring your own interests', `strongly favoring your own interests')

``Does the proverb favor autonomy or compliance?'' (`strongly favoring compliance', `favoring compliance', `indifferent', `favoring autonomy', `strongly favoring autonomy')

``Does the proverb favor trust or suspicion?'' (`strongly favoring trust', `favoring trust', `indifferent', `favoring suspicion', `strongly favoring suspicion')

``Does the proverb favor honesty or deception?'' (`strongly favoring honesty', `favoring honesty', `indifferent', `favoring deception', `strongly favoring deception')

``Does the proverb favor equality or hierarchy?'' (`strongly favoring equality', `favoring equality', `indifferent', `favoring hierarchy', `strongly favoring hierarchy')

``Does the proverb favor kindness or malice?'' (`strongly favoring kindness', `favoring kindness', `indifferent', `favoring malice', `strongly favoring malice')

``Does the proverb favor cooperation or competition?'' (`strongly favoring cooperation', `favoring cooperation', `indifferent', `favoring competition', `strongly favoring competition')

For analysis, responses were coded from 1 to 5 in the order listed above. Proverbs rated `yes' or `strong yes' on the prescriptivity question were classified as prescriptive. The analyses of the nine dimensions of judgment and behavior, including the extremity, polarity, and average recommendation results (Figure \ref{fig:polarity_preferences}), are based on this subset of prescriptive proverbs, with LLM-based results restricted by the LLM's classification and human-based results by the human raters' classification. The validation analyses (Figure \ref{fig:map}B) are based on all rated proverbs.

\subsection{Validation}

We compared the model's ratings against those of two trained research assistants, who independently rated a random sample of 500 proverbs on all ten dimensions, using the same questions and response scales as the model. To probe whether our procedure of supplying the model-generated gist affects human judgment, the raters evaluated all proverbs twice, first in a block without the model-generated gist explanation and then in a block with it.

We assessed alignment using Pearson correlations. At the level of individual proverbs, the model's ratings correlated more strongly with the average human rating than the ratings of the two human raters correlated with each other, and this held for every dimension (see Figure \ref{fig:map}B). Providing the gist explanation increased alignment both between the human raters and between humans and the model, but the pattern of results was unchanged without it. Finally, we assessed alignment at the aggregate level by correlating, across the nine dimensions, the average ratings that determine our substantive conclusions. Aggregate alignment was high for both human-model (r = .74 with gist, .69 without) and human-human comparisons (r = .84 and .80). Human raters also corroborated the model's classification of proverbs as prescriptive, judging 73.8\% of proverbs as prescriptive compared with 67.1\% according to the model. Except where results are reported separately by condition, all human results reported in this article are averages across the gist and no-gist conditions.

\subsection{Code and data availability}

The proverbs analyzed in this study comprise all proverbs listed in seven published dictionaries \parencite{mieder1992dictionary, doyle2012dictionary, mieder2021dictionary, simpson2008dictionary, knowles2016little, cordry2015multicultural, stone2006routledge}, compiled into a list of 49,528 unique proverbs. Because these works are protected by copyright, the compiled list cannot be shared publicly; it is available to editors and reviewers on request. The analysis code and the aggregated data underlying the figures are available at
\href{https://github.com/dwulff/proverb\_prescriptions}{https://github.com/dwulff/proverb\_prescriptions}. The completed GUIDE-LLM reporting checklist \parencite{feuerriegel2026reporting} is available in the same repository.

\section{Acknowledgments}
We thank Susannah Goss and Deb Ain for editing the manuscript. Dirk U. Wulff acknowledges funding by the German Research Foundation (SPP 2317, 546419617).

\subsection{Author Contributions}

Conceptualization: TL, DUW, RH; Formal analysis: DUW; Writing---original draft: TL, DUW, RH; Writing – review \& editing: TL, DUW, RH

\subsection{Competing interests}
The authors declare no competing interests.

\subsection{Use of artificial intelligence} 

The authors used Claude Opus and Claude Fable (Anthropic) to edit the manuscript. The authors reviewed all edits and take full responsibility for the content.

\printbibliography

\end{document}